\documentclass[11pt]{article}

\usepackage[preprint]{acl}

\usepackage{times}
\usepackage{latexsym}
\usepackage[T1]{fontenc}
\usepackage[utf8]{inputenc}
\usepackage{microtype}
\usepackage{graphicx}
\usepackage{booktabs}
\usepackage{amsmath}
\usepackage{amssymb}
\usepackage{subcaption}

\title{On Scope Classification and Current Knowledge-Editing Benchmarks:\\
A Negative Result, with INLAY as a Gradient-Free Case Study}

\author{Aditya Pratap Singh \\
  Independent Researcher \\
  \texttt{aditya@adityaps.work}}

\begin{document}
\maketitle

\begin{abstract}
Every memory-based knowledge editor in the SERAC lineage depends on a
\textbf{scope decision}: given a query, does a stored edit apply? We report
that current knowledge-editing benchmarks cannot measure this decision at
all. Using INLAY, a gradient-free editor we built to obtain exact per-query
ground truth (the model is frozen, edits live in an external addressable
memory, and applying an edit is a bias added along one token's unembedding
direction at decode time), we execute \emph{every} candidate router action
on $1{,}689$ queries spanning three datasets and three input conditions. An
oracle router choosing the best action every time ties a one-line static
policy to four decimal places in \emph{all nine} dataset-by-condition cells:
the maximum attainable gain of any per-query routing method is $0.00$ points.
Abstention is the sole winning action zero times out of $1{,}689$. The cause
is structural: these are counterfactual benchmarks whose evaluation question
asks for the post-edit answer, so answering from parametric knowledge is
wrong by construction, and a benchmark without negatives cannot reward a
classifier's ability to reject. This generalizes beyond our system to the
whole scope-classifier family the benchmarks are used to evaluate. We confirm
the mechanism directly: constructing the missing condition ourselves, by
withholding a query's own edit from the index for half the sample, moves
pooled headroom from exactly $+0.0000$ to $+0.0420$ and gives abstention its
first wins. We also
report where INLAY itself does not win (WISE beats it on Qwen2.5-7B
CounterFact, and retrieval-augmented generation beats every method we tested,
INLAY included, on rigorously matched RippleEdits), and disclose two bugs
found during a self-audit of our own routing machinery, neither of which
changed a published headline number outside noise.
\end{abstract}

\section{Introduction}
\label{sec:intro}

A language model's knowledge is not stored anywhere one can point to. There is
no row reading ``the CEO of $X$ is $Y$''; the fact exists as a pattern
distributed over billions of weights, each of which also participates in
thousands of unrelated facts. Correcting one fact after training (because
it changed in the world, or was wrong to begin with) is the knowledge
editing problem, and a large family of methods now addresses it by locating
weights and rewriting them \citep{meng2022rome,meng2023memit,fang2025alphaedit}
or by storing corrections outside the weights and deciding, per query, whether
a stored correction applies \citep{mitchell2022serac,hartvigsen2023grace,wang2024wise}.

That second family is the subject of this paper. Every method in it needs a
\textbf{scope classifier}: a component that looks at an incoming query and
decides whether an edit governs it, an in-context demonstration should fire,
or the model should simply answer from what it already knows. SERAC's scope
classifier \citep{mitchell2022serac} is the namesake instance, but the same
decision recurs under different names in every architecture that pairs a
frozen or lightly-modified base model with an external correction mechanism.
We built one such system, executed every action it could take on every query
in three standard benchmarks, and found that the decision this entire
architecture family is built around cannot be rewarded by the benchmarks used
to evaluate it.

\paragraph{Why we can say this with certainty rather than suspicion.} The
usual evidence for ``the benchmark doesn't test $X$'' is indirect: an ablation
that shouldn't matter doesn't, or a simpler baseline is competitive. We
instead computed the \emph{ceiling}. For $1{,}689$ queries across
CounterFact, WikiUpdate, and MQuAKE-CF in structured, unstructured, and
extracted-evidence conditions, we executed \emph{every} candidate action
(recite the stored edit, reason over retrieved evidence, or abstain) and
scored every outcome, giving per-query ground truth for which action is
correct. An oracle router that always picks the best available action is
\emph{exactly} equal to a one-line static policy in all nine cells
(\S\ref{sec:headroom}). There is no headroom for any router to find, because
abstention, the action a scope classifier exists to choose, is the sole
winning action zero times.

\paragraph{Where INLAY fits.} To get that ground truth we needed a system
whose every action could actually be executed and scored under identical
conditions, including a gradient-free recitation path with no per-edit
optimization cost. We built INLAY: the base model stays frozen, each edit is
a row in an external addressable memory (a semantic key plus the answer's
token sequence), and applying an edit adds a logit-space bias along the
answer's unembedding direction at decode time. Writing an edit is a table
insertion, about $5$--$15$\,ms with no gradient step; deleting an edit is
removing the row; and when no edit fires, output is unchanged bit for bit.
INLAY is competitive with or ahead of gradient-based editors on direct
single-fact editing (\S\ref{sec:results}), but it is not always the
strongest method we tested, and we report the cases where it loses plainly
(\S\ref{sec:results}, \S\ref{sec:ripple}). INLAY is the vehicle for this
paper's central claim, not the claim itself.

\paragraph{Contributions.}
\textbf{(i)} A negative result with a measured ceiling, not a suspicion: on
$1{,}689$ queries with per-action ground truth, an oracle router ties a
one-line static policy exactly in every cell, and abstention is never
uniquely correct (\S\ref{sec:headroom}--\S\ref{sec:cannot-measure}).
\textbf{(ii)} INLAY, a gradient-free editor with exact deletion and locality
by construction, evaluated across four model families and all three AKEW
input conditions, including evidence formats for which weight editing has no
mechanism (\S\ref{sec:method}--\S\ref{sec:results}).
\textbf{(iii)} An honest accounting of where INLAY loses: WISE beats it on
Qwen2.5-7B CounterFact, and RAG beats every method on rigorously matched
RippleEdits (\S\ref{sec:results}, \S\ref{sec:ripple}), correcting an
assumption implicit in earlier drafts of this work that the named boundary
was against weight editors, when the benchmark that actually exposes it
favors retrieval instead.
\textbf{(iv)} A transparent self-audit of our own routing machinery,
including a gate-bypass bug that made a measured mitigation inactive in the
runs behind our own published numbers (\S\ref{sec:audit}).
\textbf{(v)} A diagnosis and fix for multi-hop chain termination, more than
tripling accuracy and holding at two model scales (\S\ref{sec:multihop}).

\section{Related Work}
\label{sec:related}

\paragraph{Weight editing.}
ROME \citep{meng2022rome} treats a mid-layer MLP as a linear associative
memory and solves for a rank-one update; MEMIT \citep{meng2023memit} spreads
many updates across several layers for mass editing; AlphaEdit
\citep{fang2025alphaedit} projects the update into the null space of
preserved knowledge. All three require a clean \texttt{(prompt, target)} pair
and mutate shared parameters. \citet{hase2023localization} show that editing
at layers causal tracing does \emph{not} highlight often works equally well,
which weakens the inference from localization to edit site.

\paragraph{Memory-based and scope-classified editing.}
SERAC \citep{mitchell2022serac} introduces a scope classifier deciding
whether an edit applies, routing in-scope inputs to a counterfactual model.
MEND \citep{mitchell2022mend} learns a gradient transform. IKE
\citep{zheng2023ike} shows few-shot demonstrations can teach a model to
prefer an injected fact over its parametric belief. GRACE
\citep{hartvigsen2023grace} and WISE \citep{wang2024wise} add discrete
codebooks and side memories respectively, each with their own routing
decision at generation time. \textbf{We build directly on SERAC's
scope-classifier idea and IKE's demonstration mechanism; neither is claimed
as novel here.} Our addressing follows product-key memory
\citep{lample2019pkm} and nearest-neighbour language models
\citep{khandelwal2020knnlm}.

\paragraph{Benchmarks.}
CounterFact and zsRE evaluate single-fact edits; RippleEdits
\citep{cohen2024ripple} adds entailed consequences; MQuAKE
\citep{zhong2023mquake} adds multi-hop chains; AKEW \citep{wu2024akew}
supplies unstructured and extracted evidence alongside clean triples.
\S\ref{sec:headroom}--\S\ref{sec:cannot-measure} is a claim about what this
entire family of benchmarks can and cannot measure, independent of which
scope-classified method is being scored on them.

\begin{figure*}[t]
  \centering
  \includegraphics{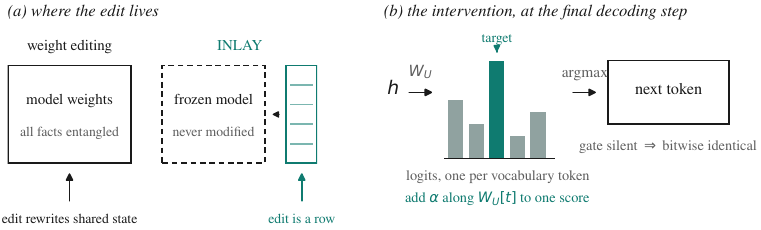}
  \caption{\textbf{(a)} Weight editing rewrites state shared by every fact.
  INLAY leaves the model frozen and appends a row to a table beside it.
  \textbf{(b)} The intervention is a bias along one token's unembedding
  direction at the final decoding step, so it moves exactly one score. When
  the gate does not fire, nothing is added and output is bitwise identical.}
  \label{fig:arch}
\end{figure*}

\section{Method: INLAY}
\label{sec:method}

INLAY exists to make the routing question in \S\ref{sec:headroom} answerable
with exact, executable ground truth: every candidate action must be a real
code path we can run and score, including a gradient-free recitation path
cheap enough to execute at scale. We describe it briefly here; it is the
vehicle for this paper's argument, not its subject.

\subsection{Design}
The model is frozen. Each edit is a row holding a key (an embedding of the
fact's question form), the answer's token sequence, and metadata
(Figure~\ref{fig:arch}a). \textbf{Writing is insertion}, not optimization:
no gradients, no covariance statistics, measured at $5$--$15$\,ms.
\textbf{Deletion is exact:} removing the row removes the edit, which matters
when removal is a regulatory obligation rather than a nicety (see the Ethics
Statement). \textbf{Edits cannot interfere:} they are physically separate
state.

\subsection{Addressing}
Keys come from a sentence encoder \citep{reimers2019sbert} rather than the
base model's hidden states: it is fast, its geometry is purpose-trained for
paraphrase matching, and it stays fixed if the base model is swapped. Keys
are compressed with a Johnson--Lindenstrauss random projection
\citep{johnson1984jl}, chosen over a learned projection because a learned one
would drift as the edit distribution changed and would silently invalidate
every stored key; a fixed random matrix never invalidates the memory, at the
cost of a few points of distortion.

\subsection{Gating and playback}
Retrieval always returns \emph{something}. A gate stacks an absolute
similarity threshold, a margin over the runner-up, and a relation-residual
check asking whether the query concerns the stored relation rather than
merely the stored subject (we return to the reliability of this gate in
\S\ref{sec:audit}). When the gate fires, playback walks the stored token
sequence and, at each decoding step, adds a bias along that token's
unembedding column so it wins the argmax (Figure~\ref{fig:arch}b). Positions
outside the answer span are untouched. Intervening in logit space rather
than hidden state is a safety argument: a logit bias moves exactly one
score, while a hidden-state edit propagates through the unembedding to every
token's logit in ways that are hard to bound; and because the intervention is
gated, a silent gate leaves behaviour bitwise unchanged.

\section{Experimental Setup}
\label{sec:setup}

Base models: GPT-2-XL (1.5B), GPT-J-6B, Qwen2.5-7B, and Mistral-7B-v0.3.
Benchmarks: CounterFact, zsRE, RippleEdits \citep{cohen2024ripple}, and AKEW
\citep{wu2024akew} over CounterFact, WikiUpdate and MQuAKE-CF in structured,
unstructured and extracted conditions. Baselines run through EasyEdit
\citep{wang2023easyedit}.

Two commitments are worth stating because they cost us results. Splits are
\textbf{subject-disjoint}: a subject never crosses a train/test boundary, so
a learned component cannot memorize an entity and appear to generalize.
Scoring uses \textbf{one convention everywhere} (diacritic- and
case-insensitive substring match against gold and its aliases) rather than
each method's internal metric, so numbers are comparable across methods that
report differently.

One implementation detail materially affects correctness: calling
\texttt{BaseEditor.edit} with \texttt{sequential\_edit=False} restores
weights before returning, so generating after it returns scores the
\emph{unedited} model.
All runs reported here use \texttt{sequential\_edit=True} with an explicit
state-dict snapshot and restore under harness control. This bug, and a
second one that specifically invalidated an earlier round of matched
RippleEdits numbers, are discussed in \S\ref{sec:audit}.

\section{Results}
\label{sec:results}

\subsection{Edit accuracy and write cost}
\label{sec:accuracy}

\begin{figure}[t]
  \centering
  \includegraphics{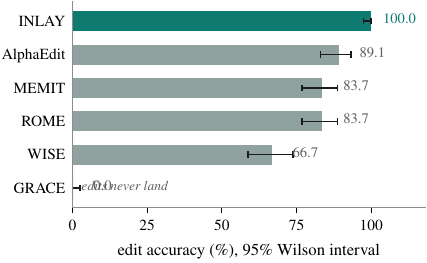}
  \caption{Edit accuracy on CounterFact, structured input, GPT-J-6B, $n=147$,
  with 95\% Wilson intervals \citep{wilson1927interval}. Scored with a single
  convention rather than each method's internal metric.}
  \label{fig:acc}
\end{figure}

Figure~\ref{fig:acc} gives the GPT-J-6B comparison. AlphaEdit is the
strongest weight editor at $89.12\%$, consistent with its null-space
constraint making each update more surgical. ROME and MEMIT tie at
$83.67\%$: single-fact CounterFact is where both converge, and MEMIT's
advantage is multi-edit preservation, not single-edit efficacy. Two results
are failures of methods that work elsewhere: WISE reaches $66.67\%$ despite
EasyEdit's own post-edit metric reporting the edit landed on $141$ of $147$
examples, because WISE routes through a side-memory module at generation
time and that routing is an additional failure point in-place updates lack;
GRACE reads exactly $0.0$ on every edit, its radius-based key rarely firing
on paraphrases. On the harmonic-mean score across ES/PS/NS at $N{=}2000$,
INLAY leads at $0.8926$ against ROME's $0.797$, WISE's $0.703$, AlphaEdit's
$0.4595$, and MEMIT's $0.4314$ (\texttt{full-results-audit.md}, Part A2).
INLAY writes in
$5$--$15$\,ms with no gradient step against seconds for gradient-based
editors, a ratio of roughly $1600\times$.

\textbf{This ranking does not hold on every model.} On Qwen2.5-7B
CounterFact at $N{=}2000$, WISE reaches a harmonic-mean score of
$\mathbf{0.9466}$ (ES $1.0$, PS $0.8553$, NS $1.0$) against INLAY's
$\mathbf{0.8944}$ at $N{=}5000$ (ES $1.0$, PS $0.8776$, NS $0.8232$):
\emph{WISE wins on Qwen}, the one place in our CounterFact results where
INLAY is not the strongest method. We report this plainly rather than
selecting the model on which INLAY leads. MEMIT and AlphaEdit's Qwen numbers
in the source logs use a different, non-comparable ``dualmetric''
probability-of-success framing rather than the harmonic-mean convention used
for INLAY/ROME/WISE/GRACE, so they are not placed in this ranking; see
Appendix~\ref{sec:appendix-tables} for both metric families side by side,
flagged as such.

\subsection{Non-structured evidence}
Weight-editing methods need a \texttt{(prompt, target)} pair to compute an
update from. They have no mechanism for ``here is a paragraph of evidence,
install whatever matters,'' so on AKEW's unstructured and extracted
conditions they are not merely worse but inapplicable; we state this as a
structural limitation rather than working around it. INLAY handles all three
conditions. Accuracy is lower on extracted input ($78.23\%$) than on
unstructured prose ($87.07\%$) despite retrieval finding the correct card
$98.64\%$ of the time in both, which locates the loss in extraction noise
rather than retrieval.

\subsection{Multi-hop}
\label{sec:multihop}

\begin{figure}[t]
  \centering
  \includegraphics{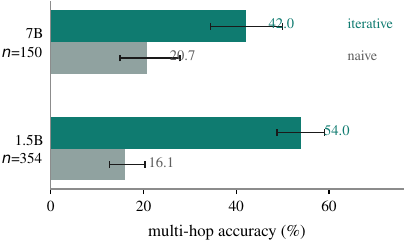}
  \caption{Iterative decomposition versus naive single-shot on MQuAKE-CF
  after the per-hop fallback fix, with 95\% Wilson intervals.}
  \label{fig:multihop}
\end{figure}

Our first iterative loop scored $5.0\%$ against a naive single-shot
baseline's $22.5\%$, four times \emph{worse} than doing nothing, with
every sampled failure producing no answer. The cause was an assumption:
MQuAKE-CF chains mix edited facts with ordinary unedited world facts ($277$
of $354$ groups contain exactly one edit across a two- or three-hop chain),
and the loop treated every hop as requiring retrieval against the edit
index, so a hop needing a never-edited fact terminated the chain answerless.
Treating a retrieval miss as a signal to answer that hop from parametric
knowledge and \emph{continue} raised accuracy to $47.5\%$ on the same sample,
and it holds at scale (Figure~\ref{fig:multihop}).

\subsection{Compositional propagation: RippleEdits}
\label{sec:ripple}

RippleEdits tests whether an edit's entailed consequences propagate while
unrelated same-subject facts are preserved, a compositional axis that
single-fact CounterFact and zsRE cannot probe. An early, less rigorous
comparison (base/RAG/INLAY sharing an identical query set) put INLAY's
preservation at $0.049$, the worst of the methods compared, and this number
was the basis for an earlier framing of INLAY's boundary as a loss to
weight-editing methods. \textbf{That framing does not survive a rigorous
comparison, and we correct it here.}

\begin{table}[t]
  \centering
  \small
  \begin{tabular}{@{}lrr@{}}
    \toprule
    Method & GPT-J-6B & Qwen2.5-7B \\
    \midrule
    in-context (RAG) & \textbf{0.3964} & \textbf{0.4381} \\
    WISE$^\dagger$   & 0.3425 & 0.1421 \\
    INLAY            & 0.2253 & 0.2871 \\
    AlphaEdit        & 0.1409 & 0.1683 \\
    ROME             & 0.1314 & 0.1453 \\
    base             & 0.0740 & 0.1522 \\
    \bottomrule
  \end{tabular}
  \caption{Matched-manifest RippleEdits (identical wikidata-verified
  subjects across every method, $n{=}100$ edits, generation-based scoring,
  edit-applied verified for every run). $\dagger$~WISE runs in its native
  sequential-accumulating mode; every other method is single-edit isolated
  with a per-edit weight restore. Source:
  the audit addendum (2026-08-19), \S B.}
  \label{tab:ripple}
\end{table}

Table~\ref{tab:ripple} gives the corrected, matched-manifest protocol:
identical wikidata-verified subjects for every method on both models,
generation-based scoring, and edit application independently verified rather
than assumed. \textbf{Retrieval-augmented generation leads every method we
tested on both models}: $0.3964$ on GPT-J-6B and $0.4381$ on Qwen2.5-7B,
ahead of every gradient-free and weight-editing method alike. INLAY does not
beat this baseline ($0.2253$ / $0.2871$). \textbf{The paper's honest
limitation is therefore that INLAY loses to RAG, not that it loses to the
weight-editing family}: on Qwen2.5-7B, INLAY's $0.2871$ beats every
weight/adapter editor we ran (ROME $0.1453$, AlphaEdit $0.1683$, WISE
$0.1421$), and on GPT-J-6B it beats ROME and AlphaEdit while trailing WISE's
side-memory accumulation mode. The mechanism behind INLAY's own weakness here
is unchanged from what we reported previously: scope precision on hard
negatives (same-subject, different-relation queries) is the open problem,
and it is exactly where a benchmark with negatives, unlike CounterFact or
zsRE, can see it.

\subsection{Sequential editing: naming which method collapses}
\label{sec:sequential}

An earlier draft of this work described weight editors as compounding
damage under repeated editing, as a general property of the family. That
claim is too broad. On GPT-2-XL, ROME's retention and locality both collapse
to $0.0$ by $n{=}25$ sequential edits (Part D of
\texttt{full-results-audit.md}), and the same collapse reproduces on GPT-J-6B
(\texttt{seq\_rome\_gptj.log}). \textbf{MEMIT does not exhibit this failure
mode on GPT-J-6B}: it holds retention and locality at $1.0$ through the
recorded checkpoints in the same sequential test
(\texttt{seq\_memit\_gptj.log}), consistent with MEMIT's design goal of
spreading updates across layers specifically to support mass editing.
\textbf{We narrow the claim to name ROME specifically}: weight editing is not
inherently unstable under repeated edits, and a method built for mass
editing (MEMIT) does not show the same failure a method built for surgical
single-edit updates (ROME) does.

\section{The Limits of Routing on Current Benchmarks}
\label{sec:headroom}

Every method in the scope-classified family includes a \textbf{scope
decision}: given a query, does a stored edit apply, and what should be done
about it? We built increasingly careful machinery for this decision,
measured statistically significant gains from it, and then found the gains
were not what they appeared to be.

\subsection{A second-order signal}
\label{sec:secondorder}
Gating was net-negative on MQuAKE-CF in every input condition, by up to
$15.87$ points. Two natural fixes failed diagnostically: recalibrating the
verifier threshold moved the false-fire rate from $18.33\%$ to $18.22\%$, and
raising the confidence threshold from $0.85$ to $0.97$ changed
\emph{nothing}: byte-identical decisions, because confidence on the
\emph{wrong} retrievals already exceeded $0.97$. The router was asking how
confident the verifier was, when it needed to know whether that confidence
was \emph{discriminative}. A verifier scoring $0.99$ on the top candidate and
$0.98$ on four unrelated ones is saturated, not confident, and no threshold
on a top-1 score can see that.

We trained a head predicting retrieval reliability from the \emph{shape} of
the candidate set (margins to the next best candidate, how many
candidates clear threshold, neighborhood entropy, subject diversity) on
CounterFact and WikiUpdate with MQuAKE-CF held out entirely. It reached
$0.956$ AUROC on the held-out dataset; $71.5\%$ of coefficient mass sat on
margin features, and the largest single weight was $-1.43$ on the best
\emph{competing} candidate's score. Used adaptively it produced $+15.9$
points on two MQuAKE-CF cells ($p{=}0.0019$, Holm-corrected
\citep{holm1979}), with zero discordant pairs where fixed gating already
worked.

\subsection{Measuring the ceiling}
\label{sec:ceiling}

\begin{figure}[t]
  \centering
  \includegraphics{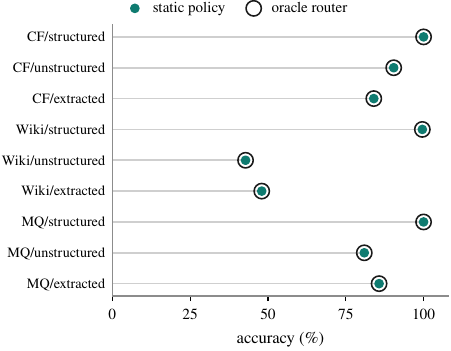}
  \caption{Every candidate action executed and scored on $1{,}689$ queries
  (CF = CounterFact, MQ = MQuAKE-CF). The oracle ring is concentric with the
  static-policy point in all nine cells: the best attainable router and one
  line of code are indistinguishable.}
  \label{fig:headroom}
\end{figure}

\begin{figure}[t]
  \centering
  \includegraphics{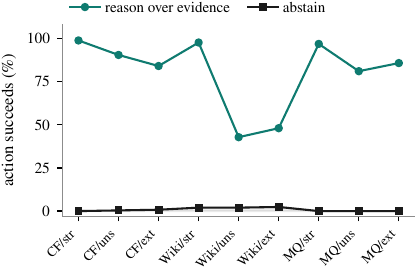}
  \caption{Per-action success rates. Abstention lies against the floor in
  every cell, which is the mechanism behind the zero headroom in
  Figure~\ref{fig:headroom}.}
  \label{fig:actions}
\end{figure}

Rather than continue improving the router, we measured its ceiling. For
$1{,}689$ queries across all three datasets and conditions we executed
\emph{every} candidate action and scored the result, yielding per-query
ground truth for which actions work.

\textbf{An oracle router choosing the best available action on every query is
exactly equal to a one-line static policy} (recite directly where legal,
otherwise reason over the retrieved evidence) to four decimal places in
all nine cells (Figure~\ref{fig:headroom}, Table~\ref{tab:headroom}). The
maximum attainable gain of any per-query routing method over one line of
code is $0.00$ points.

\begin{table}[t]
  \centering
  \small
  \begin{tabular}{@{}llrrr@{}}
    \toprule
    Dataset & Condition & $n$ & Static & Oracle \\
    \midrule
    CounterFact & structured   & 250 & 1.0000 & 1.0000 \\
    CounterFact & unstructured & 250 & 0.9040 & 0.9040 \\
    CounterFact & extracted    & 250 & 0.8400 & 0.8400 \\
    WikiUpdate  & structured   & 250 & 0.9960 & 0.9960 \\
    WikiUpdate  & unstructured & 250 & 0.4280 & 0.4280 \\
    WikiUpdate  & extracted    & 250 & 0.4800 & 0.4800 \\
    MQuAKE-CF   & structured   &  63 & 1.0000 & 1.0000 \\
    MQuAKE-CF   & unstructured &  63 & 0.8095 & 0.8095 \\
    MQuAKE-CF   & extracted    &  63 & 0.8571 & 0.8571 \\
    \midrule
    \multicolumn{2}{@{}l}{\textbf{Pooled}} & \textbf{1689}
      & \textbf{0.7874} & \textbf{0.7874} \\
    \bottomrule
  \end{tabular}
  \caption{Headroom is $+0.0000$ in every cell. ``Static'' is one line of
  code with no model and no training; ``Oracle'' is an upper bound no real
  router can exceed.}
  \label{tab:headroom}
\end{table}

The mechanism is stark (Figure~\ref{fig:actions}). Abstention succeeded on
$19$ of $1{,}689$ queries, and on all $19$ of those, reasoning or direct
recitation succeeded too. \textbf{Abstention is the sole winning action zero
times.}

\subsection{What this does to the preceding subsection}
The gains in \S\ref{sec:secondorder} are real and we stand behind the numbers. The
\emph{explanation} was wrong. We framed the head as adapting per query to
the local regime; it does not, because there is no per-query decision worth
making. What it does is suppress gates that misfire: it converges toward
the static policy. Its gains measure the harm the fixed gates were doing,
not insight it adds. A one-line rule matches it everywhere, with no training
data and none of the $k$-fold cross-encoder cost.

\section{Why Abstention Cannot Be Measured Here}
\label{sec:cannot-measure}

The reason is structural. These are \textbf{counterfactual} benchmarks: the
evaluation question asks for the \emph{post-edit} answer, so answering from
parametric knowledge returns the \emph{pre-edit} value, which is wrong by
construction. No query in the suite has ``no edit applies, answer from what
you know'' as its correct behaviour.

This generalizes beyond our system. Scope classification is load-bearing for
the SERAC-descendant family \citep{mitchell2022serac}, whose premise is a
learned decision about whether a stored edit applies. On these benchmarks
that decision is degenerate: the answer is always ``yes.'' \textbf{A
benchmark without negatives cannot measure a classifier's ability to
reject}, and an abstention path evaluated on one can only lose points.

It also explains the RippleEdits result of \S\ref{sec:ripple}. RippleEdits
\emph{does} contain same-subject, different-relation queries and therefore
partial negatives, and it is exactly there that INLAY's preservation weakness
shows up and every method's aggregate score drops relative to CounterFact and
zsRE. Scope precision matters exactly where a benchmark has negatives, which
is precisely where the main single-fact benchmarks are silent.

The missing condition is constructible: ask evaluation questions against an
index that does \emph{not} contain the corresponding edit. On such queries
abstention becomes correct, and the oracle should separate from the static
policy.

\paragraph{We built this condition, and the mechanism confirms.} For each of
the same nine cells, we removed a query's own edit card from the retrieval
index for half the sample (assigned independently per query, $\text{seed}=0$),
so REJECT is the objectively correct action for that half; the executor,
verifier, and \S\ref{sec:headroom}'s oracle-versus-static analysis are
otherwise unmodified. Pooled headroom moves from $+0.0000$ (every cell) to
$\mathbf{+0.0420}$, and REJECT-only-correct queries rise from $0/1689$ to
$52/1689$ ($3.08\%$). Splitting by population isolates the effect cleanly:
the untouched population reproduces the original near-zero result
($+0.0025$), while the edit-removed population alone carries essentially all
of the new headroom ($+0.0785$). This is a synthetic negative (a card
deleted from an otherwise intact index, not a curated out-of-scope question),
so it establishes the mechanism rather than a deployment-realistic
estimate of routing's value; we release the code
(\texttt{akew\_outcome\_labels\_oos.py}) and invite the community to build the
curated version this result motivates.

\section{Auditing Our Own Machinery}
\label{sec:audit}

A negative result about benchmark design is only as credible as the honesty
of the system used to produce it. We audited INLAY's own gating machinery
after submission-quality numbers were already in hand and report two
findings that could have changed those numbers, together with what we did to
check whether they did.

\paragraph{The margin gate is a no-op in the protocol behind our headline
numbers.} The margin-over-runner-up gate is a genuine, measured locality fix
at scale in a multi-slot regime: on a $400$-edit sequential test on
GPT-2-XL, sweeping the margin from $0$ to $0.15$ moves locality from
$0.5667$ to $1.0$ (score $0.7221 \to 0.9975$). But every headline
CounterFact and zsRE number in \S\ref{sec:accuracy} is produced under a
single-edit-per-example protocol, where memory is cleared before every
write and only one slot is ever occupied. In that regime the margin gate has
nothing to compete against and is a no-op; it only matters in multi-slot
regimes like the sequential test that measured it. We state this so the
margin gate's benefit is not read into results it did not affect.

\paragraph{A gate-bypass bug meant the relation gate was never active in the
runs behind our published matched-RippleEdits numbers.} A code audit found
that \texttt{answer\_playback()} in \texttt{gpt2\_memory\_semkey.py}, the
function every real generation path runs through, including RippleEdits,
checked only the absolute similarity score and silently skipped the margin
and relation-residual gates that the scoring-only code path already
supported. The relation gate's measured over-firing reduction (over-fire
rate $0.9667 \to 0.6667$ at \texttt{rel\_gate=0.2}) was therefore validated
only on a separate teacher-forced harness and was \textbf{never actually
active} in the matched-RippleEdits run that produced the published
preservation numbers in Table~\ref{tab:ripple}. We fixed this by adding a
single shared routing decision used identically by both the scoring and
generation code paths, then reran the exact operating point behind the
published claim with the fix genuinely active:

\begin{table}[t]
  \centering
  \small
  \begin{tabular}{@{}lrrr@{}}
    \toprule
    Aggregate & Published (bug) & Corrected & $\Delta$ \\
    \midrule
    GPT-J-6B    & 0.2253 & 0.2283 & $+0.0030$ \\
    Qwen2.5-7B  & 0.2871 & 0.2891 & $+0.0020$ \\
    \bottomrule
  \end{tabular}
  \caption{Validation rerun at \texttt{rel\_gate=0.2}, INLAY only, matched
  manifest, with the gate-bypass bug fixed. Source:
  the audit addendum (2026-08-19), \S D2.}
  \label{tab:gatebug}
\end{table}

Both deltas are within noise. \textbf{No correction to the published
headline numbers in Table~\ref{tab:ripple} was needed}, but we disclose the
bug and the check rather than silently leaving it unmentioned: the honest
reading is that the bug was real, the relation gate's isolated over-firing
metric ($0.97 \to 0.67$) does not translate into a materially stronger
matched-RippleEdits preservation score at this operating point, and the
compositional-preservation weakness \S\ref{sec:ripple} already names as an
open limitation is not solved by activating the gate here. A separate,
unrelated capacity bug in the product-key allocator (fixed on the same audit
pass) is verified not to have corrupted any published number, because the
single-edit-isolated protocol never approached the affected capacity ceiling;
see the source addendum for the full accounting.

\section{Conclusion}
\label{sec:conclusion}

The result we did not expect to report is that a one-line static policy is
provably optimal on all $1{,}689$ queries we tested, because abstention can
never be correct when every question in a counterfactual benchmark targets
an edited fact. This is a claim about the benchmarks, not about any one
system: the entire SERAC-descendant architecture family is evaluated on
suites that structurally cannot reward its central component, and reported
routing gains on them measure how much harm a fixed gate was doing rather
than any decision quality. The remedy is not a better router but a benchmark
containing queries for which the correct answer is ``this edit does not
apply.''

INLAY, the system we built to obtain this result, is independently useful:
freezing the weights and placing edits in an external addressable memory
buys writes roughly $1600\times$ cheaper than gradient-based editing, exact
deletion, and locality by construction, at the cost of reciting facts rather
than reasoning with them. It is not universally best (WISE wins on
Qwen2.5-7B CounterFact, and RAG wins on rigorously matched RippleEdits),
and we report both without qualification, alongside a self-audit that found
and checked two bugs in our own gating machinery rather than presenting
numbers uncritically.

\paragraph{Reproducibility.} All numbers come from logged runs. Splits are
subject-disjoint with seed $0$; the held-out dataset is never touched during
training of any learned component. Code, per-experiment write-ups including
negative results, and the figure generator are released with this paper.

\section*{Limitations}

\textbf{Recitation, not reasoning.} Logit-space playback reproduces a stored
answer; it cannot combine that answer with anything else, because by the
final decoding step no reasoning remains to influence. This is structural to
INLAY specifically, not to the paper's central claim about benchmark
coverage.

\textbf{Scope precision on hard negatives} is INLAY's real open problem, and
Table~\ref{tab:ripple} is the sourced evidence for it.

\textbf{Small cells.} MQuAKE-CF slices are $n{=}63$. A $+6.4$-point
structured-mode result rests on four discordant pairs, where the smallest
attainable exact McNemar $p$ is $0.125$: suggestive, not established.

\textbf{One retrieval stack.} A single encoder and one cross-encoder verifier
throughout; sensitivity to those choices is untested.

\textbf{A scale anomaly we diagnose but do not fix.} WikiUpdate is $4.38$
points \emph{worse} at 7B than at 1.5B. It is not an editing failure: the
larger model refuses to answer far more often on noisy evidence ($46.88\%$
vs $29.38\%$; $91.11\%$ vs $66.67\%$ where retrieval is wrong), and
substring accuracy scores a refusal identically to a wrong answer. The
larger model is more accurate whenever it commits; it simply commits
less. This is a property of the metric, and it implies scale comparisons on
noisy-retrieval benchmarks systematically understate larger models.

\textbf{The missing condition is only a synthetic proxy.} \S\ref{sec:cannot-measure}
constructs the missing condition by deleting a query's own edit card from an
otherwise intact index, which recovers measurable headroom ($+0.0420$ pooled)
and confirms the mechanism. This is not a curated, deployment-realistic
out-of-scope benchmark: it says nothing about how real out-of-scope
queries (never-edited facts, adversarial paraphrases, genuinely unrelated
questions) would be distributed or how much headroom they would expose. Our
claim about current benchmarks is unaffected by this caveat; the claim about
what a corrected benchmark would show is limited to what this synthetic
construction establishes.

\textbf{Coverage gaps in our own comparison table.} No MEMIT, WISE, GRACE, or
AlphaEdit results exist for Mistral-7B on CounterFact beyond the gap-fill
runs reported in Appendix~\ref{sec:appendix-tables}, and neither MEMIT nor
GRACE has been run through the matched-manifest RippleEdits protocol on
either model (Table~\ref{tab:ripple} covers base, RAG, ROME, WISE, AlphaEdit,
and INLAY only). These are open, not silently assumed favorable.

\section*{Ethics Statement}

\textbf{Deletion and the right to be forgotten.} The central structural
property of INLAY is that removing an edit is removing a row: the correction
is genuinely and verifiably gone, with no residual trace in model weights.
Gradient-based editing methods do not have this property: once a fact is
installed by a weight update, removing it again (``unlearning'') is an open
research problem, and current unlearning methods offer no comparable
guarantee of completeness. This distinction is directly relevant to
right-to-be-forgotten and GDPR-style deletion requirements for deployed
language models: an external, addressable memory architecture is a
plausible technical substrate for a system that must support real
attestable deletion, in a way that direct weight editing currently is not.
We surface this as a potential positive application, not as a claim that
INLAY as built meets any specific regulatory standard.

\textbf{Dual use: memory as an injection surface.} The same property that
makes deletion easy makes injection easy in the other direction. An external
memory that can silently override a frozen model's output at decode time is
also a mechanism for injecting misinformation into a deployed system, if the
memory store itself is not access-controlled. Unlike a weight update, which
requires retraining infrastructure and typically leaves an audit trail, a
write to an addressable memory store is fast and, absent deliberate
logging, easy to make invisible. Any deployment of this architecture should
treat the memory store itself as a security-critical component with the
same access controls one would apply to the model weights it sits beside.

\textbf{Locality is conditional on gate correctness.} We describe locality
as holding ``by construction'' when the gate does not fire, and this is true
bit-for-bit in that case. It is not a guarantee that the \emph{answer} the
gate returns is correct: locality-by-construction protects unrelated
queries from a firing gate, but a query that should be answered from
parametric knowledge and is misrouted to a stale or incorrect stored answer
will confidently return that wrong answer with no signal to the user that
anything is amiss. \S\ref{sec:ripple} and \S\ref{sec:audit} quantify how
often this misrouting occurs in our own system; a deployment should not read
``locality by construction'' as ``correctness by construction.''

\section*{Acknowledgements}

Per ACL's policy on generative AI, we disclose that an AI coding assistant
(an LLM-based agent operating under direct human supervision) was used
throughout this project: iterating on experiment code and harnesses,
running and debugging the GPU jobs behind the logged results cited
throughout this paper, drafting portions of the prose in this manuscript,
and assisting with the code audit reported in \S\ref{sec:audit}. All
research ideas, hypotheses, experimental design choices, interpretation of
results, and the decision of what to report (including the negative result
this paper is built around) are the human author's; the assistant did not
originate the paper's scientific claims. Every number in this paper traces
to a logged run inspected by the human author, and all six regression tests
released with this paper were re-run and confirmed passing before
submission. See the Responsible NLP Research checklist for the corresponding
disclosure.

\bibliography{refs}

\appendix

\section{Full Edit-Accuracy Tables}
\label{sec:appendix-tables}

Complete ES (edit success) / PS (paraphrase success) / NS (neighborhood
specificity, i.e.\ locality) / harmonic-mean-score / per-edit write-cost
tables for CounterFact and zsRE, across all four model families we ran.
Source: \texttt{outputs/full-results-audit.md}, Parts A and B, and
the audit addendum (2026-08-19), \S C.
Dashes mark a metric not reported for that run in the source logs.

\subsection{CounterFact}

Table~\ref{tab:appendix-cf} reports the complete CounterFact results across all
four model families, including the runs whose headline numbers appear in
Section~\ref{sec:results}.

\begin{table*}[t]
  \centering
  \small
  \begin{tabular}{@{}lrrrrr@{}}
    \toprule
    Method & ES & PS & NS & Score & Write (s) \\
    \midrule
    \multicolumn{6}{@{}l}{\textit{GPT-2-XL, $n=100$ (single-edit final leaderboard)}} \\
    base       & 0.00 & --   & 1.00  & 0.00 & 0.0 \\
    RAG        & 1.00 & --   & 1.00  & 1.00 & 0.0 \\
    finetune   & 1.00 & --   & 0.25  & --   & 34.7 \\
    ROME       & 0.20 & --   & 0.00  & --   & 15.0 \\
    MEMIT      & 0.40 & --   & 0.917 & --   & 10728.5$^{a}$ \\
    INLAY      & 1.00 & --   & 1.00  & --   & 0.152 \\
    \midrule
    \multicolumn{6}{@{}l}{\textit{GPT-J-6B, $N=2000$ unless noted}} \\
    base ($N{=}1000$)   & 0.004 & 0.003 & 1.00  & 0.0051 & --   \\
    RAG ($N{=}1000$)    & 0.855 & 0.464 & 0.4608 & 0.546 & --  \\
    ROME       & 0.988 & 0.7568 & 0.699 & 0.797  & 8.60 \\
    MEMIT      & 0.458 & 0.2672 & 0.972 & 0.4314 & 26.30 \\
    WISE       & 1.00  & 0.4415 & 0.999 & 0.7032 & 14.76 \\
    GRACE      & 0.00  & 0.006  & 1.00  & 0.00   & 17.35 \\
    AlphaEdit  & 0.4682 & 0.2963 & 0.982 & 0.4595 & 12.12 \\
    INLAY ($N{=}5000$) & 0.9976 & 0.8711 & 0.826 & 0.8926 & load 147.6 \\
    \midrule
    \multicolumn{6}{@{}l}{\textit{Qwen2.5-7B, $N=2000$ unless noted}} \\
    base ($N{=}5000$) & 0.0106 & 0.0108 & 1.00 & 0.016 & -- \\
    RAG ($N{=}5000$)  & 0.7983 & 0.4933 & 0.4122 & 0.5258 & -- \\
    ROME       & 0.997 & 0.5538 & 0.9487 & 0.7767 & 9.30 \\
    WISE       & 1.00  & 0.8553 & 1.00   & \textbf{0.9466} & 27.1 \\
    GRACE      & 0.00  & 0.0055 & 1.00   & 0.00   & 6.8--7.0 \\
    INLAY ($N{=}5000$) & 1.00 & 0.8776 & 0.8232 & 0.8944 & load 7.87 \\
    MEMIT$^{b}$      & \multicolumn{4}{c}{eff\_prob 1.0 / par\_prob 0.981 (dualmetric only)} & 34.3 \\
    AlphaEdit$^{b}$  & \multicolumn{4}{c}{eff\_prob 1.0 / par\_prob 0.984 (dualmetric only)} & 16.2 \\
    \midrule
    \multicolumn{6}{@{}l}{\textit{Mistral-7B-v0.3, $N=2000$ unless noted}} \\
    ROME       & 0.2846 & 0.2364 & 0.9211 & 0.3398 & 9.04 \\
    GRACE$^{d}$ & 0.00   & 0.0059 & 1.00   & 0.00   & 6.54 \\
    MEMIT$^{c}$ & 0.4218 & 0.2509 & 0.9466 & 0.4047 & -- \\
    AlphaEdit$^{d}$ & 0.4156 & 0.261 & 0.9095 & 0.4089 & 10.29 \\
    WISE$^{c}$  & 0.9962 & 0.3097 & 1.00   & 0.5733 & 10.2 \\
    INLAY ($N{=}5000$) & 1.00 & 0.8895 & 0.8257 & \textbf{0.8995} & -- \\
    \bottomrule
  \end{tabular}
  \caption{CounterFact, all model families. $^{a}$Dominated by a one-time
  3-hour covariance precompute (cached); real edit cost is seconds.
  $^{b}$MEMIT/AlphaEdit Qwen numbers exist only in the ``dualmetric''
  prob-success framing, not the EasyEdit-native harmonic-mean column used for
  every other row; not directly comparable to INLAY's $0.8944$.
  $^{c}$Mistral gap-fill runs, addendum \S C.
  $^{d}$Mistral GRACE/AlphaEdit gap-fill, closing the last open cells in the
  four-model leaderboard (\texttt{outputs/akew\_mistral\_gapfill\_results.md}).
  GRACE lands at exactly $0.0$ for the third model family in a row
  (GPT-J, Qwen, Mistral); AlphaEdit's magnitude is consistent with its
  GPT-J ($0.4595$) and Qwen (dualmetric-only) numbers.}
  \label{tab:appendix-cf}
\end{table*}

\subsection{zsRE}

Table~\ref{tab:appendix-zsre} reports the corresponding zsRE results, under the
same metrics and the same four model families.

\begin{table*}[t]
  \centering
  \small
  \begin{tabular}{@{}lrrrrr@{}}
    \toprule
    Method & ES & PS & NS & Score & Write (s) \\
    \midrule
    \multicolumn{6}{@{}l}{\textit{GPT-2-XL, $n=100$}} \\
    base       & 0.2058 & 0.2063 & 1.00 & 0.2801 & 0.0 \\
    RAG        & 0.9383 & 0.8472 & 0.88 & 0.8869 & 0.0 \\
    ROME       & 1.00 & 0.795 & 0.9564 & 0.9082 & 3.78 \\
    MEMIT      & 0.2808 & 0.2436 & 0.9961 & 0.346 & 4.37 \\
    INLAY (v3) & 0.96  & 0.96   & 1.00   & \textbf{0.973} & 0.34 \\
    \midrule
    \multicolumn{6}{@{}l}{\textit{GPT-J-6B, $N=2000$}} \\
    base       & 0.2248 & 0.2146 & 1.00 & 0.2968 & 0.0 \\
    RAG        & 0.9069 & 0.8375 & 0.8855 & 0.8757 & 0.0 \\
    ROME       & 0.9951 & 0.9412 & 0.8816 & 0.937  & 10.23 \\
    WISE       & 0.9987 & 0.9892 & 1.00   & 0.9959 & 17.28 \\
    GRACE      & 0.0126 & 0.0013 & 1.00   & 0.0035 & 18.8--18.9 \\
    MEMIT      & 0.6146 & 0.5037 & 0.9951 & 0.6497 & 26.29 \\
    AlphaEdit  & 0.6345 & 0.5265 & 0.9966 & 0.6698 & 12.21 \\
    INLAY      & 1.00 & 1.00 & 1.00 & \textbf{1.00} & 14.20 \\
    \midrule
    \multicolumn{6}{@{}l}{\textit{Qwen2.5-7B, $N=2000$}} \\
    base       & 0.3089 & 0.3045 & 1.00 & 0.3989 & 0.0 \\
    RAG        & 0.9318 & 0.8554 & 0.83 & 0.8703 & 0.0 \\
    ROME       & 0.9903 & 0.953  & 0.9828 & 0.9751 & 10.58 \\
    WISE       & 0.9991 & 0.9971 & 0.9998 & 0.9987 & 31.12 \\
    GRACE      & 0.0504 & 0.0301 & 1.00   & 0.0555 & 8.99 \\
    MEMIT      & 0.6113 & 0.57   & 0.9925 & 0.6822 & 34.80 \\
    AlphaEdit  & 0.6133 & 0.5689 & 0.9922 & 0.6824 & 15.32 \\
    INLAY      & 0.9942--0.9999 & 0.9949--0.9999 & 1.00 & 0.9963--0.9999 & 13.5--13.7 \\
    \midrule
    \multicolumn{6}{@{}l}{\textit{Mistral-7B-v0.3, $N=2000$}} \\
    ROME       & 0.6306 & 0.5972 & 0.9856 & 0.7018 & 12.50 \\
    GRACE$^{b}$ & 0.1214 & 0.0032 & 1.00   & 0.0092 & 9.11 \\
    MEMIT$^{a}$ & 0.6434 & 0.6178 & 0.9935 & 0.7178 & 19.6 \\
    AlphaEdit$^{b}$ & 0.616  & 0.5928 & 0.9913 & 0.6946 & 11.72 \\
    WISE$^{a}$  & 0.9481 & 0.9258 & 1.00   & 0.9570 & 12.7 \\
    INLAY      & 0.9999 & 0.9999 & 1.00 & \textbf{0.9999} & 13.69 \\
    \bottomrule
  \end{tabular}
  \caption{zsRE, all model families. $^{a}$Mistral gap-fill runs, addendum
  \S C.
  $^{b}$Mistral GRACE/AlphaEdit gap-fill
  (\texttt{outputs/akew\_mistral\_gapfill\_results.md}); note the ranking is
  MEMIT $>$ ROME $>$ AlphaEdit here, unlike CounterFact where ROME trails
  everything but GRACE.}
  \label{tab:appendix-zsre}
\end{table*}

\subsection{Matched-manifest RippleEdits (detail)}

Table~\ref{tab:appendix-ripple-detail} separates the propagation and
preservation components behind the matched-manifest RippleEdits aggregates
reported in Table~\ref{tab:ripple}.

\begin{table*}[t]
  \centering
  \small
  \begin{tabular}{@{}llrr@{}}
    \toprule
    Model & Method & Propagation & Preservation \\
    \midrule
    GPT-J-6B   & base      & 0.0258 & 0.1704 \\
    GPT-J-6B   & RAG       & 0.4566 & 0.2761 \\
    GPT-J-6B   & ROME      & 0.1655 & 0.0633 \\
    GPT-J-6B   & WISE      & 0.4774 & 0.0725 \\
    GPT-J-6B   & AlphaEdit & 0.1704 & 0.0819 \\
    GPT-J-6B   & INLAY     & 0.3126 & 0.0507 \\
    \midrule
    Qwen2.5-7B & base      & 0.0425 & 0.3716 \\
    Qwen2.5-7B & RAG       & 0.4533 & 0.4076 \\
    Qwen2.5-7B & ROME      & 0.1646 & 0.1067 \\
    Qwen2.5-7B & WISE      & 0.0505 & 0.3254 \\
    Qwen2.5-7B & AlphaEdit & 0.2060 & 0.0928 \\
    Qwen2.5-7B & INLAY     & 0.3458 & 0.1699 \\
    \bottomrule
  \end{tabular}
  \caption{Propagation/preservation detail behind the aggregates in
  Table~\ref{tab:ripple}. Source:
  the audit addendum (2026-08-19), \S B.}
  \label{tab:appendix-ripple-detail}
\end{table*}

\end{document}